\documentclass[•]{amsart}

\usepackage{graphicx,amsmath,amssymb,booktabs,tabulary,multirow,overpic,xcolor}
\definecolor{citecolor}{RGB}{34,139,34}
\usepackage[pagebackref=true,breaklinks=true,letterpaper=true,colorlinks,
  citecolor=citecolor,bookmarks=false]{hyperref}
\usepackage[british,english,american]{babel}

\title{Adapting Mask-RCNN for Automatic Nucleus Segmentation}
\author{Jeremiah W. Johnson}
\address{University of New Hampshire, Manchester, NH 03101}
\email{jeremiah.johnson@unh.edu}

\begin{document}

\begin{abstract} 

Automatic segmentation of microscopy images is an important task in medical image processing and analysis. Nucleus detection is an important example of this task. Mask-RCNN is a recently proposed state-of-the-art algorithm for object detection, object localization, and object instance segmentation of natural images. In this paper we demonstrate that Mask-RCNN can be used to perform highly effective and efficient automatic segmentations of a wide range of microscopy images of cell nuclei, for a variety of cells acquired under a variety of conditions.   

\end{abstract}

\maketitle

%%%%%%%%%%%%%%%%%%%%%%%%%%%%%%%%%%%%%%%%%%%%%%%%%%%%%%%%%%%%%%%%%%%%%%%%%
\section{Introduction} 
%%%%%%%%%%%%%%%%%%%%%%%%%%%%%%%%%%%%%%%%%%%%%%%%%%%%%%%%%%%%%%%%%%%%%%%%%

In the last few years, algorithms based on convolutional neural networks (CNNs) have led to dramatic advances in the state of the art for fundamental problems in computer vision, such as object detection, object localization, semantic segmentation, and object instance segmentation. \cite{Krizhevsky:2012:ICD:2999134.2999257}, \cite{DBLP:journals/corr/SimonyanZ14a}, \cite{resnets}, \cite{DBLP:journals/corr/ZagoruykoLLPGCD16}. This has led to increased interest in the applicability of convolutional neural network-based methods for problems in medical image analysis. Recent work has shown promising results on tasks as diverse as automated diagnosis diabetic retinopathy, automatic diagnosis of melanoma, precise measurement of a patient's cardiovascular ejection fraction, segmentation of liver and tumor 3D volumes, segmentation of mammogram images, and 3D knee cartilage segmentation \cite{Pratt2016200}, \cite{esteva_skin_cancer}, \cite{Prasoon2013}, \cite{DBLP:journals/corr/Tran16}, \cite{liver_segmentation}, \cite{mammogram}, \cite{ejection_fraction}.

Semantic segmentation of natural images is a long-standing and not fully solved computer vision problem, and in the past few years progress in this area has been almost exclusively driven by CNN-based models. Notable developments in recent years include the development of RCNN in 2014, fully convolutional neural networks in 2015, and the development in 2015 of the Fast-RCNN followed by the Faster-RCNN models \cite{girshick14CVPR}, \cite{fully_convolutional}, \cite{fast_rcnn}, \cite{faster_rcnn}. These algorithms were designed with semantic segmentation, object localization, and object instance segmentation of natural images in mind. Recently there has been an explosion of development in this area, with convolutional neural network based architectures such as Feature Pyramid Networks, SegNets, RefineNets, DilatedNets, and Retinanets developed all pushing benchmarks for this task forward \cite{DBLP:journals/corr/LinDGHHB16}, \cite{badrinarayanan2015segnet}, \cite{Lin:2017:RefineNet}, \cite{DBLP:journals/corr/abs-1708-02002}, \cite{dilatednet}. In addition, single-shot models such as YOLO and SSD have enabled object detection to occur at speeds up to 100-1000 times faster than region proposal based algorithms \cite{yolov3}, \cite{ssd}.

In 2015, the U-Net architecture was developed explicitly with the segmentation of medical images in mind, and used to produce state-of-the-art results on on the ISBI challenge for segmentation of neuronal structures in electron microscopic stacks as well as the ISBI cell tracking challenge 2015 \cite{unet}. U-Net architectures have since been adapted and used for a wide range of tasks in medical image analysis including volumetric segmentation of 3D structures and sparse-view CT reconstructions \cite{Han2017FramingUV}, \cite{3d_unet}.  
 
%%%%%%%%%%%%%%%%%%%%%%%%%%%%%%%%%%%%%%%%%%%%%%%%%%%%%%%%%%%%%%%%
\subsection{The Mask-RCNN Model} 
 %%%%%%%%%%%%%%%%%%%%%%%%%%%%%%%%%%%%%%%%%%%%%%%%%%%%%%%%%%%%%%%
 
\begin{figure}[t]\label{mask_rcnn_image}

\begin{center}
	\includegraphics[scale=0.5]{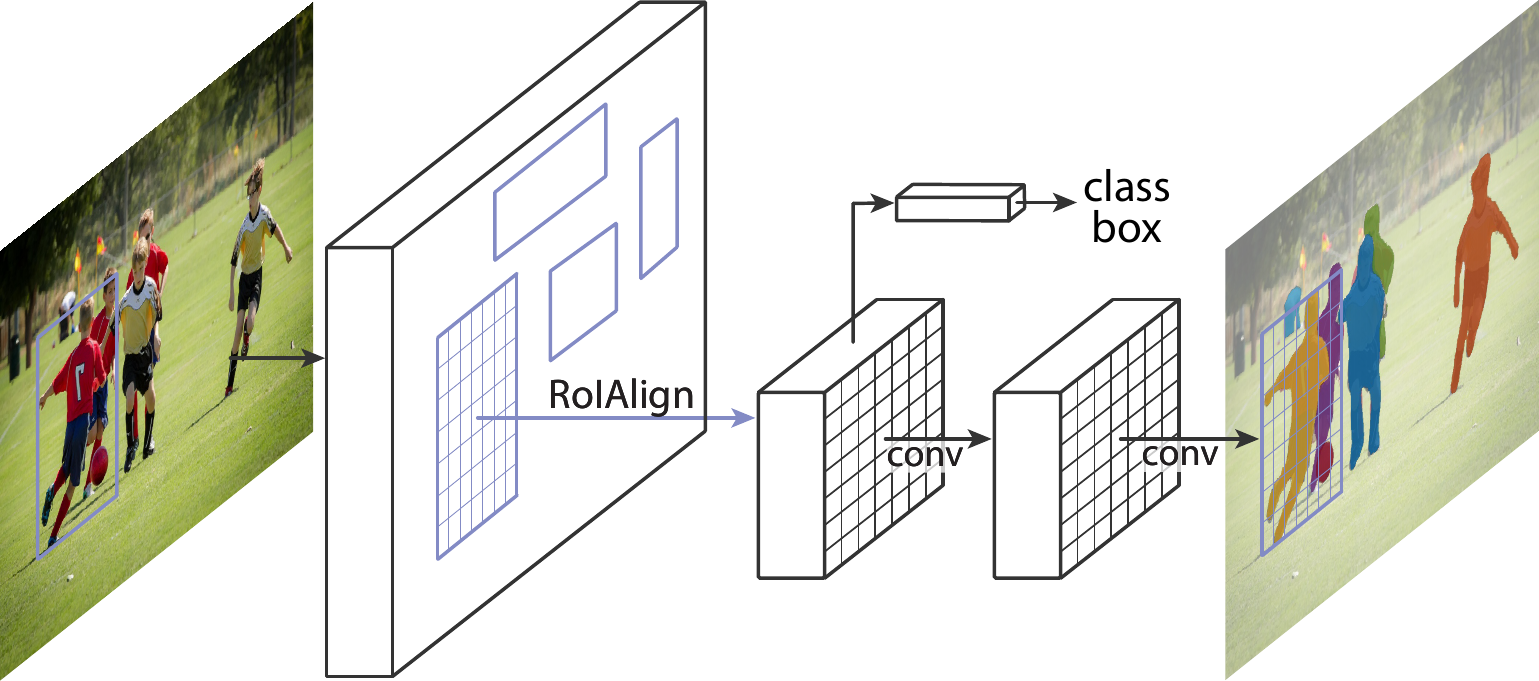}
\end{center}

\caption{The Mask-RCNN model. Image from \cite{mask_rcnn}. Used with permission.}
\end{figure}
 
The Mask-RCNN model was developed in 2017 and extends the Faster-RCNN model for semantic segmentation, object localization, and object instance segmentation of natural images \cite{mask_rcnn}. Mask-RCNN is described by the authors as providing a `simple, flexible and general framework for object instance segmentation'. Mask-RCNN was used to outperform all existing single-model entries on every task in the 2016 COCO Challenge, a large-scale object detection, segmentation, and captioning challenge \cite{mscoco}. 

Many modern algorithms for image segmentation fall into one of two classes: those that rely on a region proposal algorithm and those that do not. U-Net, for instance, is an example of a segmentation algorithm that does not rely on a region proposal algorithm; rather, U-Net uses an encoder-decoder framework in which a convolutional neural network learns, or encodes, a representation of the content of the image and a second network, such as a deconvolutional neural network, constructs the desired segmention mask from the learned representation produced by the encoder (note that a deconvolutional neural network may also be referred to as a fully convolutional neural network, a transposed convolutional neural network or a fractionally-strided convolutional neural network in the literature) \cite{fully_convolutional}, \cite{johnson2016perceptual}. Encoder-decoder architectures have been used in machine learning for a variety of tasks outside of object detection or segmentation for some time, such as denoising images or generating images \cite{kingma_vae}, \cite{denoising_autoencoders}.

Mask-RCNN, in contrast, relies on a region proposals which are generated via a region proposal network. Mask-RCNN follows the Faster-RCNN model of a feature extractor followed by this region proposal network, followed by an operation known as ROI-Pooling to produce standard-sized outputs suitable for input to a classifier, with three important modifications. First, Mask-RCNN replaces the somewhat imprecise ROI-Pooling operation used in Faster-RCNN with an operation called ROI-Align that allows very accurate instance segmentation masks to be constructed; and second, Mask-RCNN adds a network head (a small fully convolutional neural network) to produce the desired instance segmentations; c.f. Figure \ref{mask_rcnn_image}. Finally, mask and class predictions are decoupled; the mask network head predicts the mask independently from the network head predicting the class. This entails the use of a multitask loss function $L = L_{cls} + L_{bbox} + L_{mask}$. For additional details, we refer interested readers to \cite{mask_rcnn}.

Mask-RCNN is built on a backbone convolutional neural network architecture for feature extraction \cite{girshick14CVPR}, \cite{faster_rcnn}. In principle, the backbone network could be any convolutional neural network designed for images analysis, such as ResNet-50 or ResNet-101 \cite{resnets}; however, it has been shown that using a feature pyramid network (FPN) based on a network such as ResNet-50 or ResNet-101 as the Mask-RCNN backbone gives gains in both accuracy and speed \cite{mask_rcnn}. A feature pyramid network takes advantage of the inherent hierarchical and multi-scale nature of convolutional neural networks to derive useful features for object detection, semantic segmentation, and instance segmentation at many different scales. Feature pyramid network models all require a `backbone` network themselves in order for the feature pyramid to be constructed.  Again, the backbone model is typically chosen to be a convolutional neural network known for high performance at object detection, and may be pretrained \cite{DBLP:journals/corr/LinDGHHB16}.   

Although the properties of natural images will in general differ significantly from medical images, given the effectiveness of Mask-RCNN at general-purpose object instance segmentation, it is a reasonable candidate for use in automated segmentation of medical images. Here, we investigate the efficacy of a Mask-RCNN model at detecting nuclei in microscopy images.

%%%%%%%%%%%%%%%%%%%%%%%%%%%%%%%%%%%%%%%%%%%%%%%%%%%%%%%%%%%%%%%%%%%%%%%%
% Sample Images and Masks
%%%%%%%%%%%%%%%%%%%%%%%%%%%%%%%%%%%%%%%%%%%%%%%%%%%%%%%%%%%%%%%%%%%%%%%%

\begin{figure}[tp]\label{sample_images}

  \centering

  \label{figure}

  \begin{tabular}{ccc}

    \includegraphics[width=30mm]{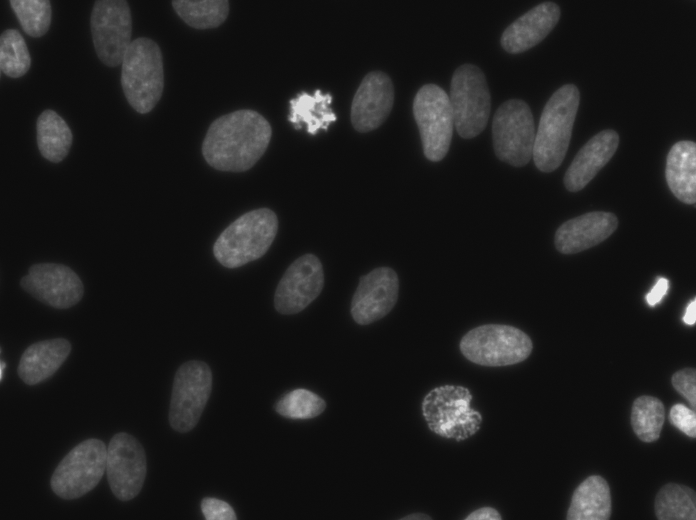} &

    \includegraphics[width=30mm]{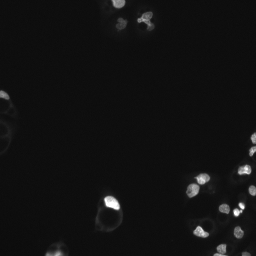} &
    
    \includegraphics[width=30mm]{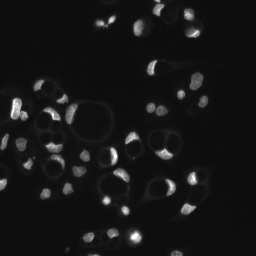} \\

    \includegraphics[width=30mm]{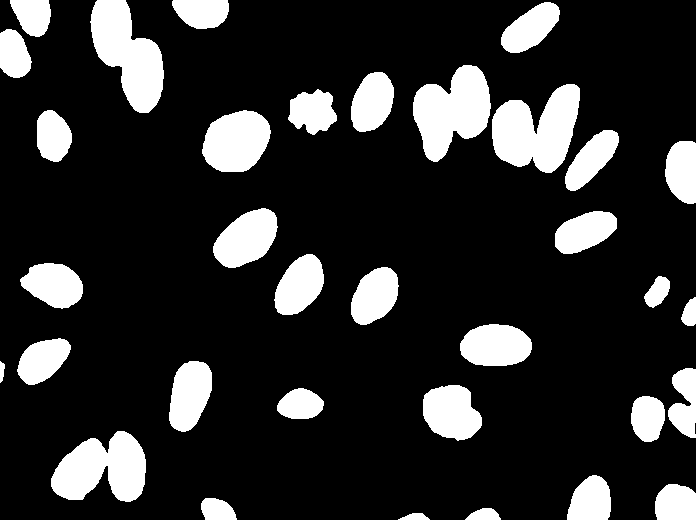} &

    \includegraphics[width=30mm]{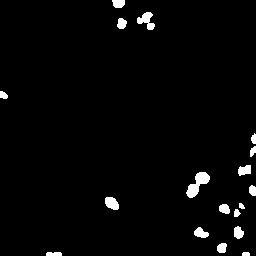} &

    \includegraphics[width=30mm]{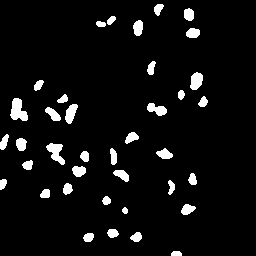} \\

  \end{tabular}
  \caption{Sample Nuclei Images and Masks. For each image, an individual mask is provided for each nucleus detected. To generate the masks in the bottom row, all of the masks provided for each image have been merged into a single mask. Note that images vary widely, including in size.}

\end{figure}

%%%%%%%%%%%%%%%%%%%%%%%%%%%%%%%%%%%%%%%%%%%%%%%%%%%%%%%%%%%%%%%%%%%%%%%%%
\section{The Data}
%%%%%%%%%%%%%%%%%%%%%%%%%%%%%%%%%%%%%%%%%%%%%%%%%%%%%%%%%%%%%%%%%%%%%%%%%

The data used for these experiments is image set \href{https://data.broadinstitute.org/bbbc/BBBC038/}{BBBC038v1}, available from the Broad Bioimage Benchmark Collection \cite{nature_broad}. These data were used for Stage 1 of the the 2018 Data Science Bowl, an annual competition sponsored by Booz Allen Hamilton and hosted on the data science website \href{http://www.kaggle.com}{kaggle.com}. The data consist of 729 microscopy images and corresponding annotations for each individual nucleus detected by an expert in each image; c.f. Figure \ref{sample_images}. The nuclei in the images are derived from a wide range of organisms including humans, mice, and flies. Furthermore, the nuclei in the images have been imaged and treated in variety of conditions and appear in a variety of contexts and states, including tissues and embryos, and cell division and genotoxic stress. This presents a significant additional challenge as convolutional neural networks can be expected to perform best in general when the input data is as uniform and standardized as possible. This includes standardization in terms of color, contrast, scale, and class balance. 

Of these 729 images in the dataset, 664 images were used for training and validating the model and 65 images were held out for testing.

%%%%%%%%%%%%%%%%%%%%%%%%%%%%%%%%%%%%%%%%%%%%%%%%%%%%%%%%%%%%%%%%%%%%%%%%%
\section{Methodology}
%%%%%%%%%%%%%%%%%%%%%%%%%%%%%%%%%%%%%%%%%%%%%%%%%%%%%%%%%%%%%%%%%%%%%%%%%

For all experiments described here, we use a Mask-RCNN model with a feature pyramid network backbone. The implementation used is based on an existing implementation by Matterport Inc. released under an MIT License, and which is itself based on the open-source libraries Keras and Tensorflow \cite{matterport}, \cite{tensorflow}, \cite{keras}. This implementation is well-documented and easy to extend. For these experiments, we tried both a ResNet-50 feature pyramid network model and a ResNet-101 feature pyramid network model as a backbone. We note that the model with ResNet-50-FPN backbone has a somewhat lower computational load than that with a ResNet-101 backbone, but the ResNet-101-FPN gives significantly improved results with no other changes to the model or training procedure. Rather than training the network end-to-end from the start, we initialize the model using weights obtained from pretraining on the MSCOCO dataset \cite{mscoco} and proceed to train the layers in three stages: first, training only the network heads, which are randomly initialized, then training the upper layers of the network (from stage 4 and up in the ResNet model), and then reducing the learning rate by a factor of 10 and training end to end. In total we train for 100 epochs using stochastic gradient descent with momentum of 0.9, starting with a learning rate of 0.001 and ending with a learning rate of 0.0001. Although we experimented with longer and shorter training times, additional training did not lead to noticable improvement and few epochs led to underfit. We use a batch size of 6 on a single NVIDIA Titan Xp GPU. Gradients are clipped to 5.0 and weights are decayed by 0.0001 each epoch. We also conducted additional experiments with other learning rate schedules, but they gave no additional improvements and aren't reported here.

For all of these experiments, image preprocessing is kept to a minimum. Images are upsampled by a factor of two and the channel means are normalized. We conducted multiple experiments on mirroring the image edges by various numbers of pixels to improve detection of small nuclei at the edges as described in \cite{unet}, but we saw no improvement from doing so and omitted this step from the final algorithm.To help avoid overfitting, the dataset was augmented using random crops, random rotations, gaussian blurring, and random horizontal and vertical flips.

%%%%%%%%%%%%%%%%%%%%%%%%%%%%%%%%%%%%%%%%%%%%%%%%%%%%%%%%%%%%%%%%%%%%%%%%%%
% Results Table
%%%%%%%%%%%%%%%%%%%%%%%%%%%%%%%%%%%%%%%%%%%%%%%%%%%%%%%%%%%%%%%%%%%%%%%%%%

\begin{table}[t]
\begin{tabular}{l|l|c}
Backbone       &    AP                &  Mask Average IoU \\ \hline
ResNet-50 FPN  &  56.06               &      66.98        \\
ResNet-100 FPN &  59.40               &      70.54
\end{tabular}\vspace{2mm}

\caption{\textbf{Instance segmentation} mask results on validation data. All results are single-model results.}
\label{results_table}\vspace{-3mm}
\end{table}

\subsection{Results}
\begin{figure}[tp]\label{sample_detections}

  \centering

  \label{figure}

  \begin{tabular}{ccc}

    \includegraphics[width=40mm]{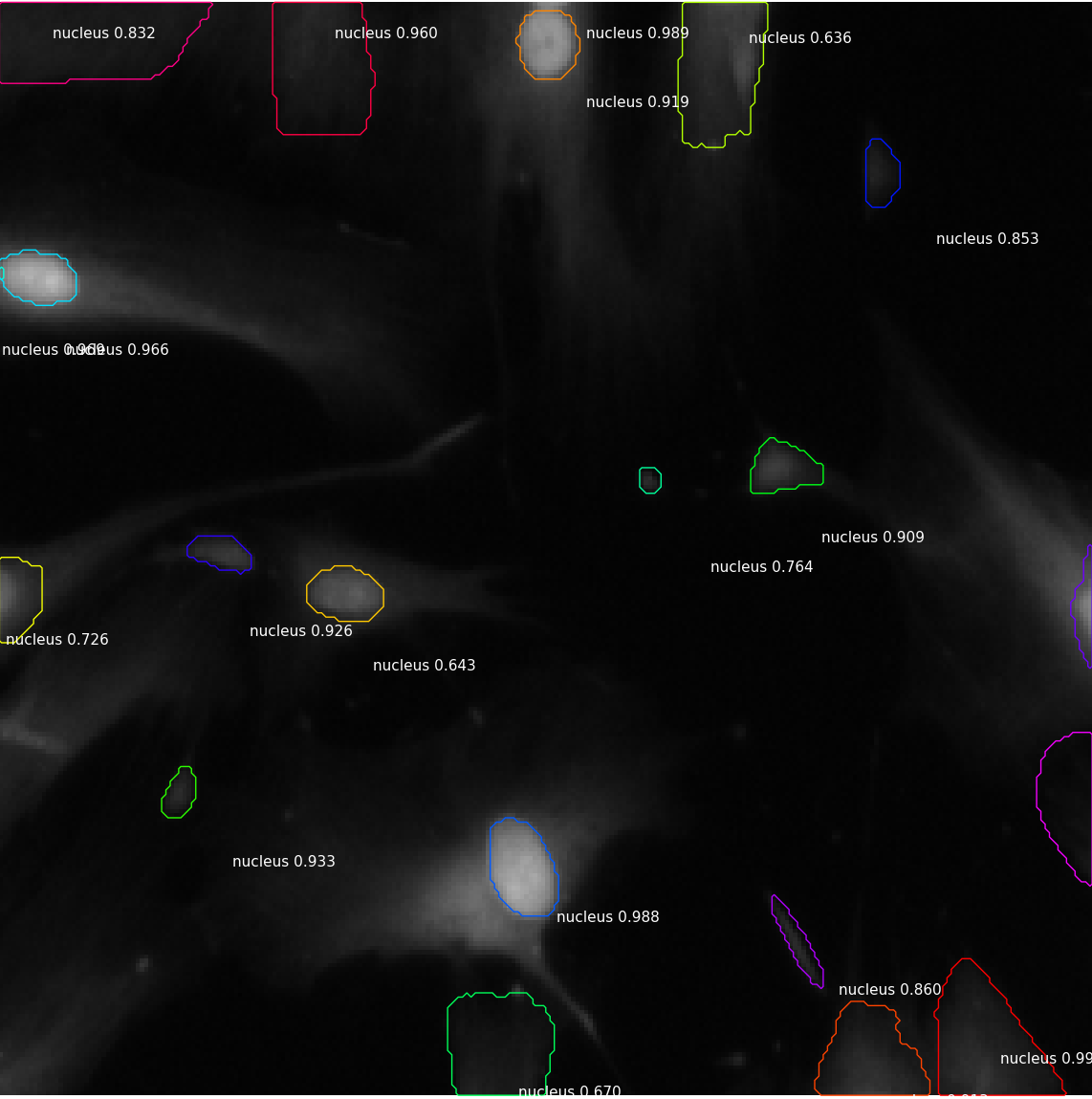} &

    \includegraphics[width=40mm]{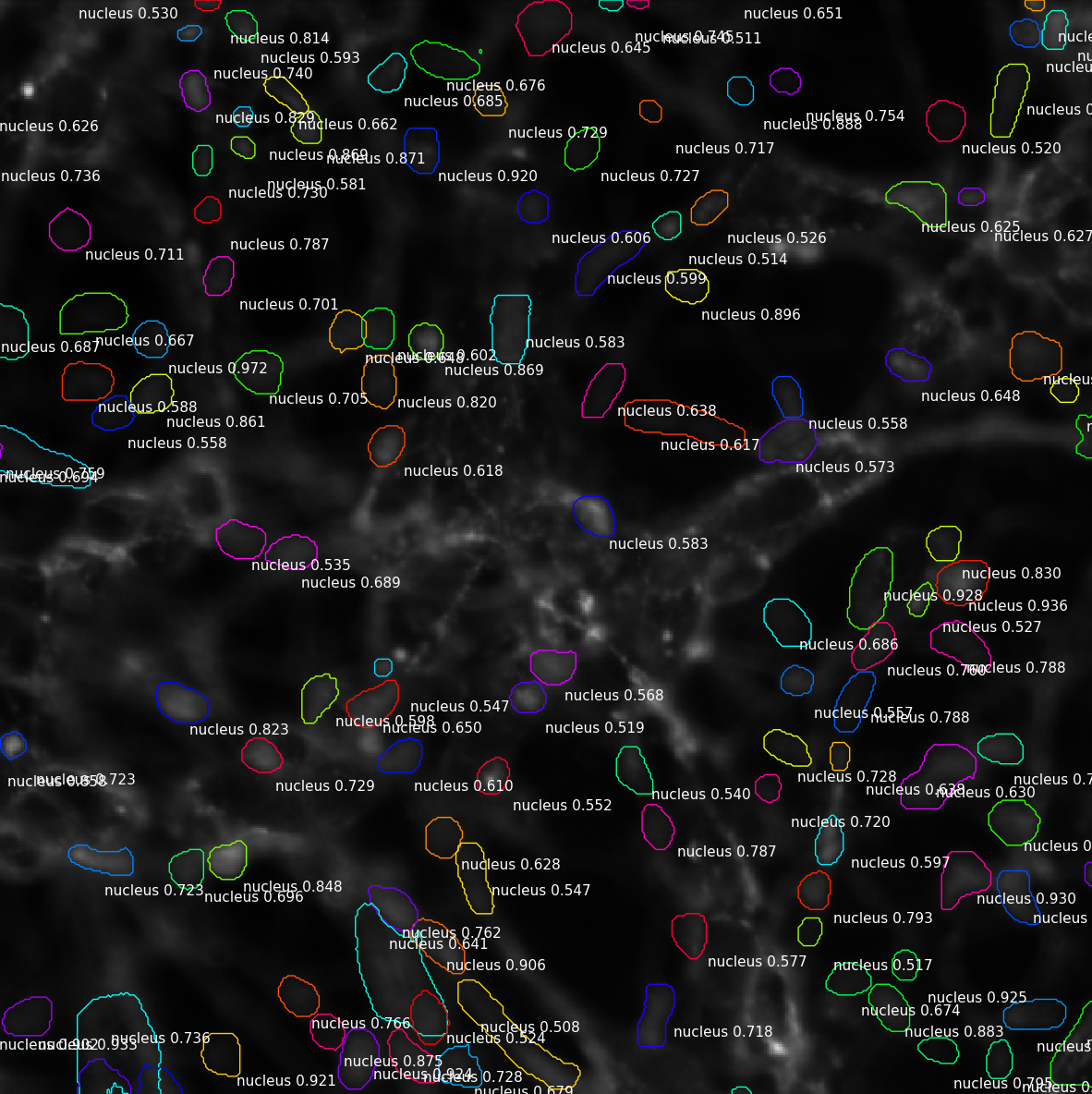} &
    
    \includegraphics[width=40mm]{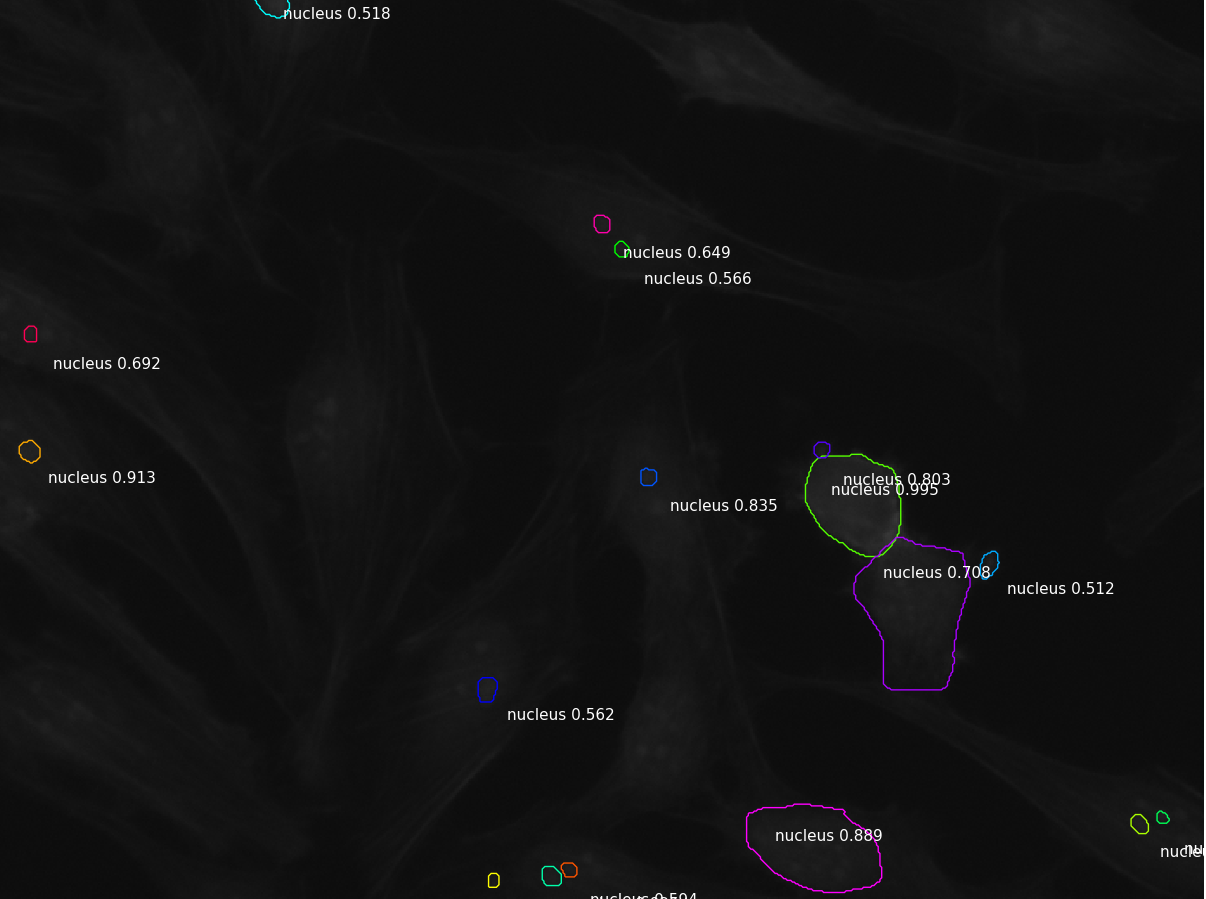} 

  \end{tabular}
  \caption{Sample detections from the ResNet-50-FPN model.}

\end{figure}

The model with ResNet-50 backbone and parameters as described above obtains an average mask intersection over union (IoU) of 66.98\% on the validation dataset. The mean average precision at thresholds 0.5 to 0.95 by steps of size 0.05 as defined as the primary metric for the MSCOCO challenge (AP) is 56.06\% for this model \cite{mscoco}. The mode with ResNet-101 backbone and the same parameters and training procedures as described above obtains an average mask IoU of 70.54\% and a mean average precision as defined for the MSCOCO challenge of 59.40\%. These results are summarized in table \ref{results_table}. Several sample detections are illustrated in Figure \ref{sample_detections}. 

%%%%%%%%%%%%%%%%%%%%%%%%%%%%%%%%%%%%%%%%%%%%%%%%%%%%%%%%%%%%%%%%%%%%%%%%%
% Sample Detections
%%%%%%%%%%%%%%%%%%%%%%%%%%%%%%%%%%%%%%%%%%%%%%%%%%%%%%%%%%%%%%%%%%%%%%%%%
 
%%%%%%%%%%%%%%%%%%%%%%%%%%%%%%%%%%%%%%%%%%%%%%%%%%%%%%%%%%%%%%%%%%%%%%%%%
\section{Conclusion \& Future Work}
%%%%%%%%%%%%%%%%%%%%%%%%%%%%%%%%%%%%%%%%%%%%%%%%%%%%%%%%%%%%%%%%%%%%%%%%%

In this paper we demonstrate that the Mask-RCNN model, while primarily designed with object detection, object localization, and instance segmentation of natural images in mind, can be used to produce high quality results for the challenging task of segmentation of nuclei in widely varying microscopy images with very little modification. There are several similar tasks in medical image analysis for which it is likely that a Mask-RCNN based model could easily be adapted to improve performance without extensive modification or customization. Examples of this include the task of segmentation of the left ventricle of the heart, where accurate segmentations can be used to estimate a cardiac patient's ejection fraction and improve their outcomes, or liver and tumor segmentation as described in \cite{liver_segmentation}. Future work will explore the efficacy and performance of Mask-RCNN-based models for a range of such tasks. 

%%%%%%%%%%%%%%%%%%%%%%%%%%%%%%%%%%%%%%%%%%%%%%%%%%%%%%%%%%%%%%%%%%%%%%%%%%
\subsubsection*{Acknowledgments}
%%%%%%%%%%%%%%%%%%%%%%%%%%%%%%%%%%%%%%%%%%%%%%%%%%%%%%%%%%%%%%%%%%%%%%%%%%

The author would like to thank NVIDIA Corp. for GPU donation to support this research.

%%%%%%%%%%%%%%%%%%%%%%%%%%%%%%%%%%%%%%%%%%%%%%%%%%%%%%%%%%%%%%%%%%%%%%%%%%
\bibliography{bibliography_nucleus}

\begin{thebibliography}{10}

\bibitem{matterport}
Mask-{RCNN}.
\newblock \url{https://github.com/matterport/Mask_RCNN}.
\newblock Accessed: 2018-04-27.

\bibitem{tensorflow}
{\sc Abadi, M., Agarwal, A., Barham, P., Brevdo, E., Chen, Z., Citro, C.,
  Corrado, G.~S., Davis, A., Dean, J., Devin, M., Ghemawat, S., Goodfellow, I.,
  Harp, A., Irving, G., Isard, M., Jia, Y., Jozefowicz, R., Kaiser, L., Kudlur,
  M., Levenberg, J., Man\'{e}, D., Monga, R., Moore, S., Murray, D., Olah, C.,
  Schuster, M., Shlens, J., Steiner, B., Sutskever, I., Talwar, K., Tucker, P.,
  Vanhoucke, V., Vasudevan, V., Vi\'{e}gas, F., Vinyals, O., Warden, P.,
  Wattenberg, M., Wicke, M., Yu, Y., and Zheng, X.}
\newblock {TensorFlow}: Large-scale machine learning on heterogeneous systems,
  2015.
\newblock Software available from tensorflow.org.

\bibitem{badrinarayanan2015segnet}
{\sc Badrinarayanan, V., Kendall, A., and Cipolla, R.}
\newblock Segnet: A deep convolutional encoder-decoder architecture for image
  segmentation.
\newblock {\em IEEE Transactions on Pattern Analysis and Machine
  Intelligence\/} (2017).

\bibitem{keras}
{\sc Chollet, F., et~al.}
\newblock Keras.
\newblock \url{https://keras.io}, 2015.

\bibitem{liver_segmentation}
{\sc Christ, P.~F., Ettlinger, F., Gr{\"{u}}n, F., Elshaer, M. E.~A.,
  Lipkov{\'{a}}, J., Schlecht, S., Ahmaddy, F., Tatavarty, S., Bickel, M.,
  Bilic, P., Rempfler, M., Hofmann, F., D'Anastasi, M., Ahmadi, S., Kaissis,
  G., Holch, J., Sommer, W.~H., Braren, R., Heinemann, V., and Menze, B.~H.}
\newblock Automatic liver and tumor segmentation of {CT} and {MRI} volumes
  using cascaded fully convolutional neural networks.
\newblock {\em CoRR abs/1702.05970\/} (2017).

\bibitem{3d_unet}
{\sc {\c C}i{\c c}ek, {\"O}., Abdulkadir, A., S.~Lienkamp, S., Brox, T., and
  Ronneberger, O.}
\newblock {\em 3D U-Net: Learning Dense Volumetric Segmentation from Sparse
  Annotation}.
\newblock 10 2016.

\bibitem{esteva_skin_cancer}
{\sc Esteva, A., Kuprel, B., Novoa, R.~A., Ko, J., Swetter, S.~M., Blau, H.~M.,
  and Thrun, S.}
\newblock Dermatologist-level classification of skin cancer with deep neural
  networks.
\newblock {\em Nature 542\/} (01 2017), 115 -- 118.

\bibitem{fast_rcnn}
{\sc Girshick, R.}
\newblock Fast {R-CNN}.
\newblock {\em arXiv preprint arXiv:1504.08083\/} (2015).

\bibitem{girshick14CVPR}
{\sc Girshick, R., Donahue, J., Darrell, T., and Malik, J.}
\newblock Rich feature hierarchies for accurate object detection and semantic
  segmentation.
\newblock In {\em Computer Vision and Pattern Recognition\/} (2014).

\bibitem{Han2017FramingUV}
{\sc Han, Y., and Ye, J.~C.}
\newblock Framing {U-Net} via deep convolutional framelets: Application to
  sparse-view {CT}.
\newblock {\em CoRR abs/1708.08333\/} (2017).

\bibitem{mask_rcnn}
{\sc He, K., Gkioxari, G., Doll{\'{a}}r, P., and Girshick, R.~B.}
\newblock Mask {R-CNN}.
\newblock {\em CoRR abs/1703.06870\/} (2017).

\bibitem{resnets}
{\sc He, K., Zhang, X., Ren, S., and Sun, J.}
\newblock Deep residual learning for image recognition.
\newblock {\em CoRR abs/1512.03385\/} (2015).

\bibitem{johnson2016perceptual}
{\sc Johnson, J., Alahi, A., and Fei-Fei, L.}
\newblock Perceptual losses for real-time style transfer and super-resolution.
\newblock In {\em European Conference on Computer Vision\/} (2016), Springer,
  pp.~694--711.

\bibitem{ejection_fraction}
{\sc Kabani, A., and El-Sakka, M.~R.}
\newblock Ejection fraction estimation using a wide convolutional neural
  network.
\newblock In {\em Image Analysis and Recognition\/} (Cham, 2017), F.~Karray,
  A.~Campilho, and F.~Cheriet, Eds., Springer International Publishing,
  pp.~87--96.

\bibitem{kingma_vae}
{\sc Kingma, D.~P., and Welling, M.}
\newblock Auto-encoding variational bayes.
\newblock {\em arXiv preprint arXiv:1312.6114\/} (2013).

\bibitem{Krizhevsky:2012:ICD:2999134.2999257}
{\sc Krizhevsky, A., Sutskever, I., and Hinton, G.~E.}
\newblock Imagenet classification with deep convolutional neural networks.
\newblock In {\em Proceedings of the 25th International Conference on Neural
  Information Processing Systems - Volume 1\/} (USA, 2012), NIPS'12, Curran
  Associates Inc., pp.~1097--1105.

\bibitem{Lin:2017:RefineNet}
{\sc Lin, G., Milan, A., Shen, C., and Reid, I.}
\newblock Refine{N}et: {M}ulti-path refinement networks for high-resolution
  semantic segmentation.
\newblock In {\em CVPR\/} (July 2017).

\bibitem{DBLP:journals/corr/LinDGHHB16}
{\sc Lin, T., Doll{\'{a}}r, P., Girshick, R.~B., He, K., Hariharan, B., and
  Belongie, S.~J.}
\newblock Feature pyramid networks for object detection.
\newblock {\em CoRR abs/1612.03144\/} (2016).

\bibitem{DBLP:journals/corr/abs-1708-02002}
{\sc Lin, T., Goyal, P., Girshick, R.~B., He, K., and Doll{\'{a}}r, P.}
\newblock Focal loss for dense object detection.
\newblock {\em CoRR abs/1708.02002\/} (2017).

\bibitem{mscoco}
{\sc Lin, T.-Y., Maire, M., Belongie, S., Hays, J., Perona, P., Ramanan, D.,
  Doll{\'a}r, P., and Zitnick, C.~L.}
\newblock Microsoft {COCO}: Common objects in context.
\newblock In {\em Computer Vision -- ECCV 2014\/} (Cham, 2014), D.~Fleet,
  T.~Pajdla, B.~Schiele, and T.~Tuytelaars, Eds., Springer International
  Publishing, pp.~740--755.

\bibitem{ssd}
{\sc Liu, W., Anguelov, D., Erhan, D., Szegedy, C., Reed, S., Fu, C.-Y., and
  Berg, A.~C.}
\newblock Ssd: Single shot multibox detector.
\newblock To appear.

\bibitem{nature_broad}
{\sc Ljosa, V., Sokolnicki, K.~L., and Carpenter, A.~E.}
\newblock Annotated high-throughput microscopy image sets for validation.
\newblock {\em Nature Methods 9\/} (06 2012), 637 EP --.

\bibitem{Prasoon2013}
{\sc Prasoon, A., Petersen, K., Igel, C., Lauze, F., Dam, E., and Nielsen, M.}
\newblock {\em Deep Feature Learning for Knee Cartilage Segmentation Using a
  Triplanar Convolutional Neural Network}.
\newblock Springer Berlin Heidelberg, Berlin, Heidelberg, 2013, pp.~246--253.

\bibitem{Pratt2016200}
{\sc Pratt, H., Coenen, F., Broadbent, D.~M., Harding, S.~P., and Zheng, Y.}
\newblock Convolutional neural networks for diabetic retinopathy.
\newblock {\em Procedia Computer Science 90\/} (2016), 200 -- 205.
\newblock 20th Conference on Medical Image Understanding and Analysis (MIUA
  2016).

\bibitem{yolov3}
{\sc Redmon, J., and Farhadi, A.}
\newblock Yolov3: An incremental improvement.
\newblock {\em arXiv\/} (2018).

\bibitem{faster_rcnn}
{\sc Ren, S., He, K., Girshick, R.~B., and Sun, J.}
\newblock Faster {R-CNN:} towards real-time object detection with region
  proposal networks.
\newblock {\em CoRR abs/1506.01497\/} (2015).

\bibitem{unet}
{\sc Ronneberger, O., P.Fischer, and Brox, T.}
\newblock U-net: Convolutional networks for biomedical image segmentation.
\newblock In {\em Medical Image Computing and Computer-Assisted Intervention
  (MICCAI)\/} (2015), vol.~9351 of {\em LNCS}, Springer, pp.~234--241.
\newblock (available on arXiv:1505.04597 [cs.CV]).

\bibitem{fully_convolutional}
{\sc Shelhamer, E., Long, J., and Darrell, T.}
\newblock Fully convolutional networks for semantic segmentation.
\newblock {\em IEEE Trans. Pattern Anal. Mach. Intell. 39}, 4 (Apr. 2017),
  640--651.

\bibitem{DBLP:journals/corr/SimonyanZ14a}
{\sc Simonyan, K., and Zisserman, A.}
\newblock Very deep convolutional networks for large-scale image recognition.
\newblock {\em CoRR abs/1409.1556\/} (2014).

\bibitem{DBLP:journals/corr/Tran16}
{\sc Tran, P.~V.}
\newblock A fully convolutional neural network for cardiac segmentation in
  short-axis {MRI}.
\newblock {\em CoRR abs/1604.00494\/} (2016).

\bibitem{denoising_autoencoders}
{\sc Vincent, P., Larochelle, H., Lajoie, I., Bengio, Y., and Manzagol, P.-A.}
\newblock Stacked denoising autoencoders: Learning useful representations in a
  deep network with a local denoising criterion.
\newblock {\em Journal of Machine Learning Research 11}, Dec (2010),
  3371--3408.

\bibitem{dilatednet}
{\sc Yu, F., and Koltun, V.}
\newblock Multi-scale context aggregation by dilated convolutions.
\newblock In {\em ICLR\/} (2016).

\bibitem{DBLP:journals/corr/ZagoruykoLLPGCD16}
{\sc Zagoruyko, S., Lerer, A., Lin, T., Pinheiro, P. H.~O., Gross, S.,
  Chintala, S., and Doll{\'{a}}r, P.}
\newblock A multipath network for object detection.
\newblock {\em CoRR abs/1604.02135\/} (2016).

\bibitem{mammogram}
{\sc Zhu, W., Xiang, X., Tran, T.~D., Hager, G.~D., and Xie, X.}
\newblock Adversarial deep structured nets for mass segmentation from
  mammograms.
\newblock {\em CoRR abs/1710.09288\/} (2017).

\end{thebibliography}
%%%%%%%%%%%%%%%%%%%%%%%%%%%%%%%%%%%%%%%%%%%%%%%%%%%%%%%%%%%%%%%%%%%%%%%%%%

\bibliographystyle{acm}

\end{document}